\documentclass[10pt,twocolumn,letterpaper]{article}

\usepackage{booktabs}

\usepackage{iccv}
\usepackage{times}
\usepackage{epsfig}
\usepackage{graphicx}
\usepackage{amsmath}
\usepackage{amssymb}
\usepackage{bm}
\usepackage{nidanfloat}
\usepackage{subfig}
\usepackage{dsfont}
\usepackage{enumitem}
\usepackage{algorithm}
\usepackage[noend]{algpseudocode}
\usepackage{multirow}
\usepackage{makecell}
\newcolumntype{H}{>{\setbox0=\hbox\bgroup}c<{\egroup}@{}}
\usepackage{arydshln}
\usepackage{epstopdf}
\usepackage{eso-pic}
\usepackage{framed}
\newenvironment{nscenter}
 {\parskip=0pt\par\nopagebreak\centering}
 {\par\noindent\ignorespacesafterend}

\algnewcommand{\LineComment}[1]{\State \# {#1}}

\usepackage[pagebackref=true,breaklinks=true,letterpaper=true,colorlinks,bookmarks=false]{hyperref}

\iccvfinalcopy 

 \AddToShipoutPicture*{\footnotesize \sffamily\raisebox{1.5cm}{\hspace{2.4cm}
\begin{minipage}{15cm}
DOI:  10.1109/ICCV48922.2021.00511 \quad © 2021 IEEE. Personal use of this material is permitted. Permission from IEEE must be
obtained for all other uses, in any current or future media, including
reprinting/republishing this material for advertising or promotional purposes, creating new
collective works, for resale or redistribution to servers or lists, or reuse of any copyrighted
component of this work in other works.
\end{minipage}
 }}

\begin{document}

\title{BlockCopy:  High-Resolution Video Processing with \\Block-Sparse Feature Propagation and Online Policies}

\author{  {Thomas Verelst \quad Tinne Tuytelaars}\\
{ESAT-PSI, KU Leuven}\\
{Leuven, Belgium}\\
{\tt\small \{thomas.verelst, tinne.tuytelaars\}@esat.kuleuven.be}
}

\maketitle
\ificcvfinal\thispagestyle{empty}\fi

\begin{abstract}
In this paper we propose BlockCopy, a scheme that accelerates pretrained frame-based CNNs to process video more efficiently, compared to standard frame-by-frame processing. To this end, a lightweight policy network determines important regions in an image, and operations are applied on selected regions only, using custom block-sparse convolutions. Features of non-selected regions are simply copied from the preceding frame, reducing the number of computations and latency. The execution policy is trained using reinforcement learning in an online fashion without requiring ground truth annotations. Our universal framework is demonstrated on dense prediction tasks such as pedestrian detection, instance segmentation and semantic segmentation, using both state of the art (Center and Scale Predictor, MGAN, SwiftNet) and standard baseline networks (Mask-RCNN, DeepLabV3+). BlockCopy achieves significant FLOPS savings and inference speedup with minimal impact on accuracy.\end{abstract}

\section{Introduction}
\begin{figure}[t!]
\begin{nscenter}
\includegraphics[width=0.9\linewidth,trim={0 0.15cm 0 0}, clip]{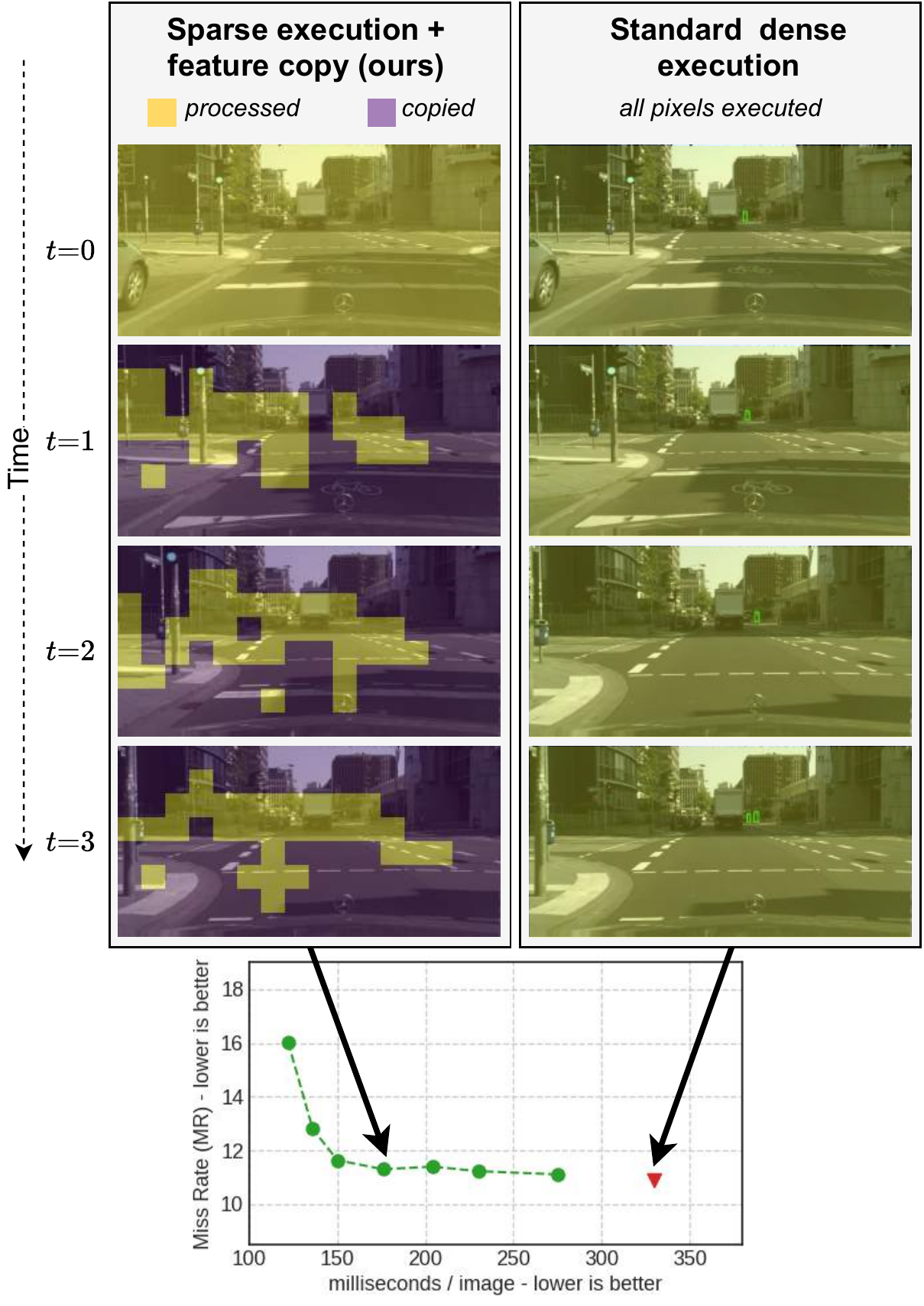}
\vspace*{-0.5em}
\caption{BlockCopy accelerates existing CNNs by sparsely executing convolutions, while copying features from previous executions in non-important regions. In this example on pedestrian detection, inference speed is more than doubled with negligible increase in detection miss rate. 
\label{fig:teaser}}
\end{nscenter}
\end{figure}

Most contemporary convolutional neural networks (CNN) are trained on images and process video frame-by-frame, for simplicity or due to the lack of large annotated video datasets. For instance, the popular COCO dataset~\cite{lin2014microsoft} for large-scale object detection does not include video sequences. However, video typically contains a considerable amount of redundancy in the temporal domain, with some image regions being almost static.
Image-based convolutional neural networks do not take advantage of temporal and spatial redundancies to improve efficiency: they apply the same operations on every pixel and every frame. Representation warping has been proposed to save computations~\cite{gadde2017semantic_netwarp, zhu2017deep_dff, jain2019accel}, but optical flow is expensive and warping cannot cope with large changes such as newly appearing objects. Other video processing methods, \eg using 3D convolutions or recurrent neural networks~\cite{ji20123d_3dconv, maturana2015voxnet_3dconv, liu2019looking_fast_and_slow}, focus on improving accuracy by using temporal information, instead of reducing computations by exploiting redundancies.

In this work, we propose a method to improve the efficiency and inference speed of convolutional neural networks for dense prediction tasks, by combining temporal feature propagation with sparse convolutions as illustrated in Figure~\ref{fig:teaser}.  A lightweight, trainable policy network selects important image regions, and the expensive task network sparsely executes convolutions on selected regions only. Features from non-important regions are simply copied from the previous execution, thereby saving computations. 

The policy network is trained with reinforcement learning in an online fashion: the output of the large task network, for example Mask-RCNN~\cite{he2017mask}, serves as supervisory signal to train the online policy. 
Using online reinforcement learning has several advantages. First, no labeled data is required and off-the-shelf networks can be optimized during deployment without designing a separate training pipeline. Second, online training allows the network to fine-tune the policy to the task and dataset at deployment time. 
Finally, models of different computational costs can be obtained by simply adjusting the policy's computational target parameter.

The main contributions of this work are as follows:
\begin{itemize}[noitemsep, topsep=0pt]
    \itemsep0em
    \item We propose BlockCopy to adapt existing CNNs for more efficient video processing, using block-sparse convolutions and temporal feature propagation. Our framework is implemented in PyTorch, using custom CUDA operations.
    \item We utilize reinforcement learning to train a policy network in an online fashion without requiring ground-truth labels. 
    \item We demonstrate our method on pedestrian detection, instance segmentation and semantic segmentation tasks and show that existing off-the-shelf CNNs can be significantly accelerated without major compromises in accuracy. 
    \item We show that BlockCopy improves the accuracy-speed trade-off by comparison with existing methods, lower resolution and lower frame rate baselines.
\end{itemize}
The code is available online\footnote{\scriptsize{\url{https://github.com/thomasverelst/blockcopy-video-processing-pytorch}}}.

\section{Related work}

Well-known methods to reduce the computational cost of convolutional neural networks are pruning~\cite{li2016pruning}, quantization~\cite{hubara2017quantized} or knowledge distillation~\cite{hinton2015distilling}. Recently, dynamic methods~\cite{ferrari_skipnet_2018, wu2018blockdrop} gained interest, adapting the network's operations based on the image's difficulty. Video processing methods are complementary, as they avoid redundant computations using the temporal dimension.

\subsection{Conditional execution and sparse processing}

Dynamic neural networks, also known as conditional execution~\cite{bengio_estimating_2013_conditional, veit2018convolutional_convnet_aig}, adapt the network complexity based on the image's difficulty. SkipNet~\cite{ferrari_skipnet_2018} and ConvNet-AIG~\cite{veit2018convolutional_convnet_aig} skip residual blocks for easy images, reducing the average amount of computations for image classification. The policy, determining which blocks to skip, is  learned using reinforcement learning~\cite{ferrari_skipnet_2018} or a reparametrization trick~\cite{veit2018convolutional_convnet_aig}. Recently, this has been extended to the spatial domain~\cite{figurnov2017spatially_sact, xie2020spatially_sampling, verelst2020dynamic} by skipping individual pixels. However, pixel-wise sparse convolutions are not supported by default in deep learning frameworks such as PyTorch and TensorFlow, and are challenging to implement efficiently on GPU~\cite{gale2020sparse}. As a consequence, most work only demonstrates performance improvements on CPU~\cite{xie2020spatially_sampling}, on specific architectures~\cite{verelst2020dynamic}, or only consider theoretical advantages~\cite{figurnov2017spatially_sact}.

Block-based processing, where pixels are grouped in blocks, are more feasible and have been accelerated on GPU~\cite{ren_sbnet_2018, verelst2020segblocks}. Simply splitting images in blocks and then processing those individually is not sufficient, as features should propagate between blocks to achieve a large receptive field and avoid boundary artefacts. To this end, \mbox{SBNet}~\cite{ren_sbnet_2018} proposes to use partly overlapping blocks and applies this for 3D object detection.
SegBlocks~\cite{verelst2020segblocks} introduces a framework with BlockPadding modules to process images in blocks. In this work, we extend conditional execution to the video domain, using a policy trained with reinforcement learning in combination with block-based sparse processing and feature transfer. 
 
\subsection{Video processing}
Most video processing methods focus on video classification and action recognition applications~\cite{diba2017temporal_videoclassification, tran2018closer_actionrecognition}, incorporating multi-frame motion information using methods such as multi-frame non-maximum suppression~\cite{han2016seq_seqnms}, feature fusion~\cite{liu2020efficient_perframe}, 3D convolutions~\cite{ji20123d_3dconv,maturana2015voxnet_3dconv}, recurrent neural networks~\cite{liu2019looking_fast_and_slow} or other custom operations~\cite{wang2018nonlocal}. 

Efficiency and speed can be improved by exploiting temporal redundancies, as changes between frames are often small. Clockwork Nets~\cite{shelhamer2016clockwork} proposed a method to adaptively execute network stages based on semantic stability.
Deep Feature Flow~\cite{zhu2017deep_dff}, NetWarp~\cite{gadde2017semantic_netwarp}, GRFP~\cite{nilsson2018semantic_grfp}, Awan and Shin~\cite{awan2020warping}, and Paul~\etal~\cite{paul2020efficient_evs} warp representations  between frames using optical flow, with Accel~\cite{jain2019accel} introducing a lightweight second branch to fine tune representations. DVSNet~\cite{xu2018dynamic_dvs} proposes an adaptive keyframe scheduler, selecting the key frames to execute while other frames use warped representations.  Awan and Shin~\cite{awan2020online} use reinforcement learning to train the keyframe selection scheme. However, optical flow is an expensive calculation and therefore these methods mainly focus on large networks such as DeepLabV3+~\cite{chen2018encoder_deeplab}, where the introduced overhead is modest compared to the segmentation network. Jain and Gonzalez~\cite{jain2018fast_bmv} use block motion vectors, already present in compressed video. Low-Latency Video Semantic Segmentation (LLVSS)~\cite{li2018low_llvss} does not use optical flow, but updates keyframe representations using a lightweight per-frame update branch. Mullapudi~\etal~\cite{mullapudi2019online_distillation} demonstrate online knowledge distillation for video object segmentation, where a lightweight student network is finetuned for a given situation in the video by learning from a large teacher network. Our method does not require keyframes, and only the first frame of each clip is executed completely. All other frames are executed sparsely, resulting in more consistent processing delays in comparison with keyframe-based methods.

\section{BlockCopy method and policy network}
BlockCopy optimizes a large task network for more efficient video processing, by combining block-sparse convolutions with feature transfers. Our method consists of two main components: a framework to efficiently execute a CNN architectures in a block-sparse fashion using temporal feature propagation, and a policy network determining whether blocks should be executed or transferred.

\begin{figure*}[tb!]
\centering

\includegraphics[width=1\linewidth]{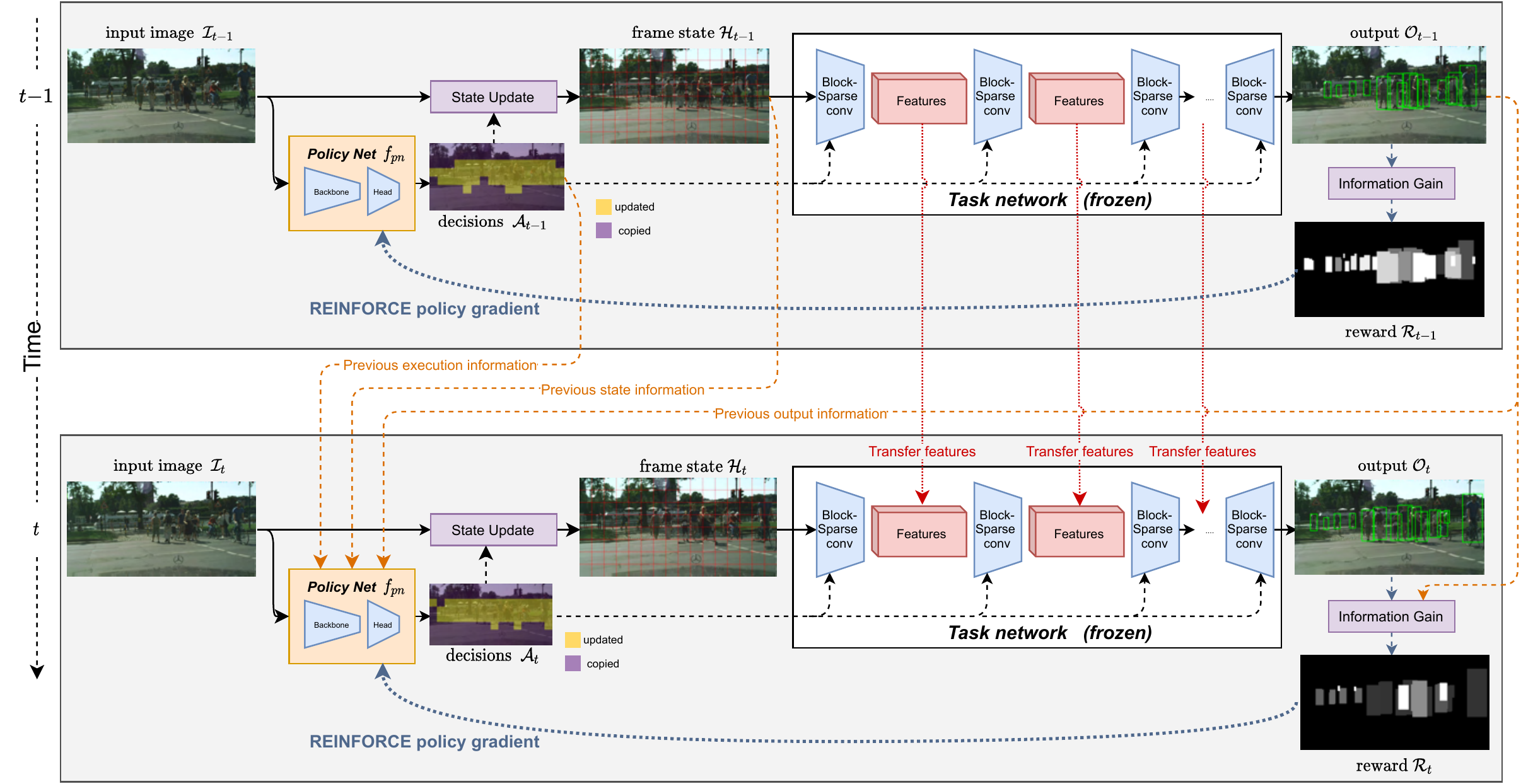}
\vspace*{-4mm}
\caption{Overview of the BlockCopy pipeline, illustrated for two video frames. The policy network outputs execution decisions. The network is executed using block-sparse convolutions and features from the previous iteration are copied to non-executed regions. The importance of each block is measured using Information Gain, which serves as a reward to update the policy weights.
\label{fig:detail_overview}}
\end{figure*}

The policy network is a lightweight, trainable convolutional network, selecting the blocks to be executed. As the decision is binary for each region, \textit{execute} or \textit{transfer}, standard backpropagation cannot be used to train the policy network. Therefore, we use reinforcement learning, based on a reward per block, to train the policy network in an online self-distilled fashion based on the task network's output. 
The reward function is based on the \textit{information gain}, representing the amount of task information gained by executing the region instead of just transferring features. 
Note that the task network's weights are not updated, and only the policy network is trained.
Figure~\ref{fig:detail_overview} presents an overview of the components discussed in the next subsections. 

\subsection{Block-sparse processing with feature transfer}

Standard libraries for deep learning such as PyTorch~\cite{paszke2019pytorch} do not support efficient sparse convolutions. We build on the framework introduced by SegBlocks~\cite{verelst2020segblocks} to process images in blocks, by first splitting images into blocks and applying their BlockPadding module avoiding discontinuities between blocks. At execution time, representations throughout the network are stored and copied with efficient and specialized CUDA modules. 
\begin{figure*}[tb!]
\centering
\includegraphics[width=1\linewidth]{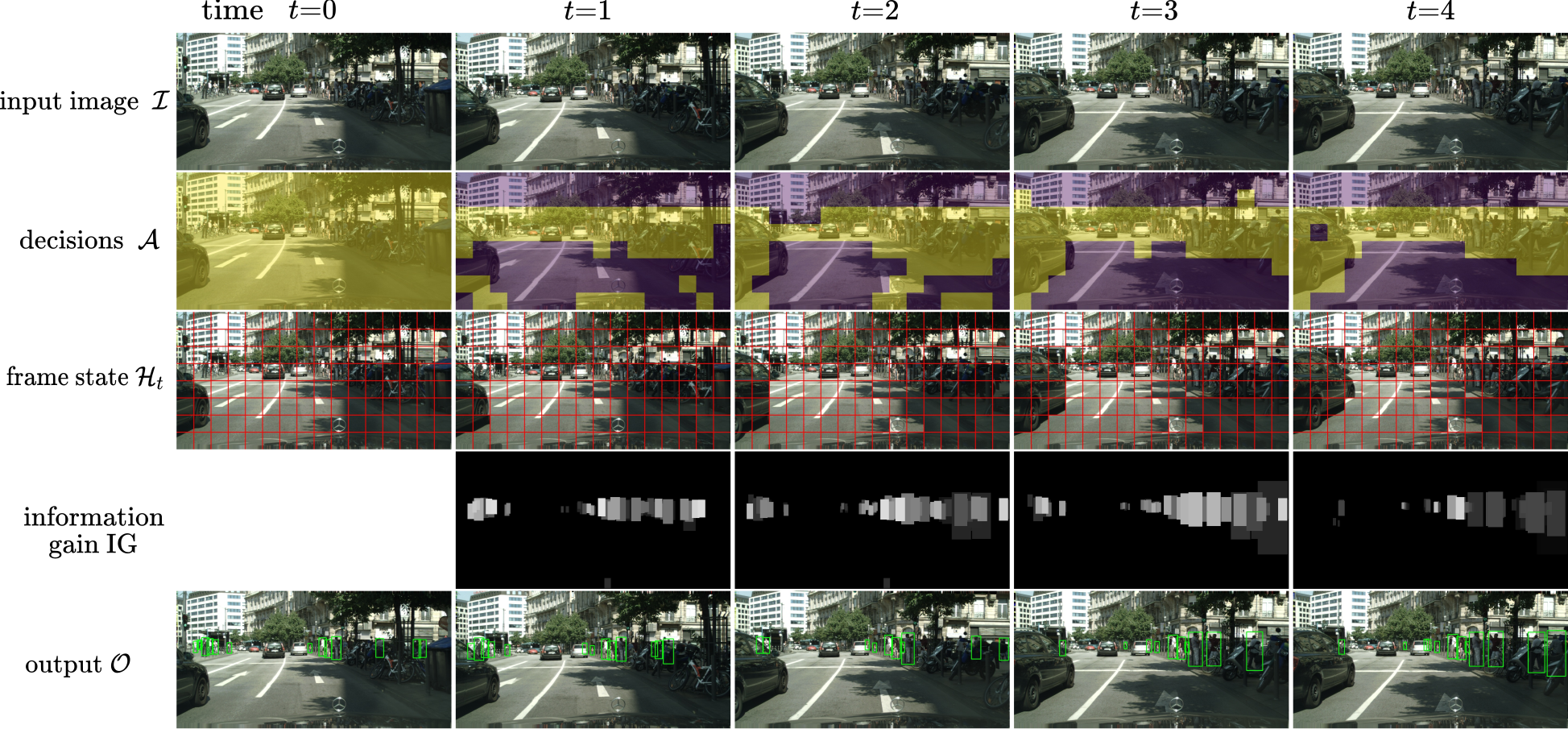}
\vspace*{-6mm}
\caption{Illustration of a video sequence, with execution grids, frame states and outputs. The frame state is only updated for selected regions (yellow), whereas features from other regions are re-used from the previous frame (purple). The output bounding boxes are visualized for detections with scores larger than 0.5, whereas the reward scales with the detection score.
A video with visualizations can be found in supplemental material. \
\label{fig:video}}
\end{figure*}

\subsection{Online policy with reinforcement learning}
The policy network is trained to select important regions that have high impact on the output. Using ground-truth annotations of video sequences, one could extract the regions where the output changes. However, many computer vision datasets do not contain video sequences and annotating ground-truth is expensive. Instead of using ground-truth annotations, we opt for a more flexible approach with self-distillation and online reinforcement learning. 

When a block is executed, the importance of this execution is determined using the \textit{information gain}. Blocks where large changes in the output occur have a large information gain. This way, the policy network can learn the relative importance of blocks at execution time, without requiring a separate training pipeline, expensive teacher network or annotated data. 

\subsection{Policy network architecture}

\begin{figure*}[tb!]
\centering
\includegraphics[width=1\linewidth]{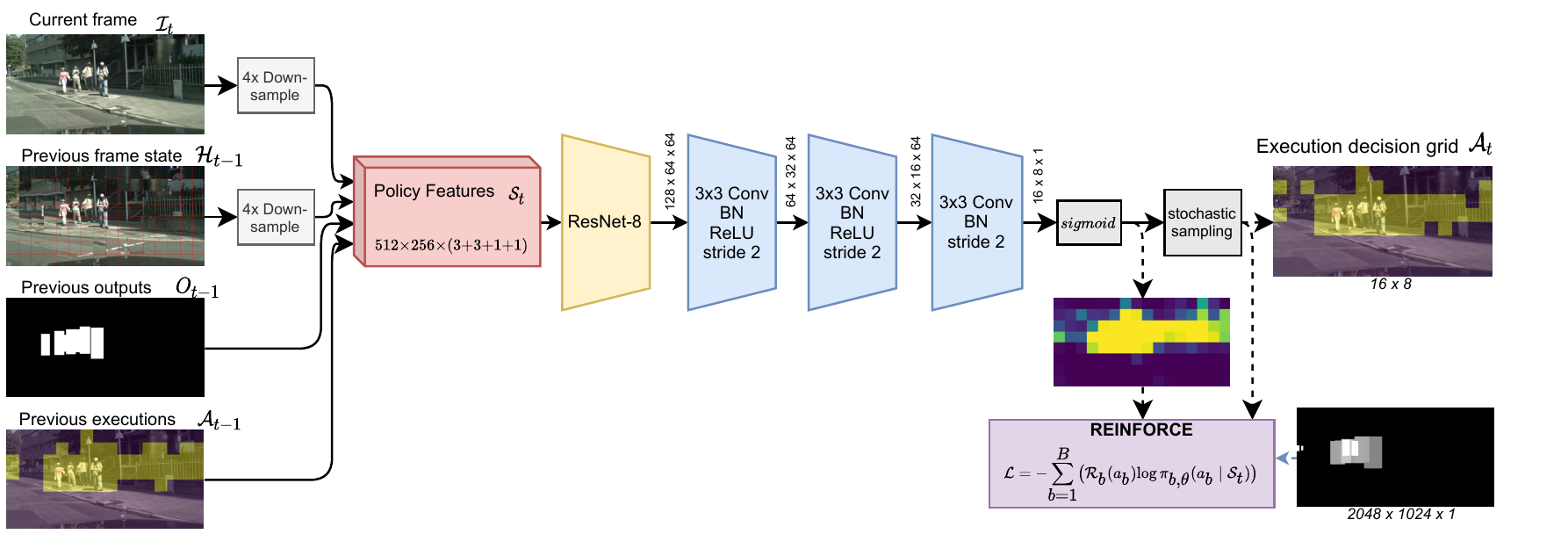}
\vspace*{-7mm}
\caption{Policy network architecture. Dimensions are given as $W{\times}H{\times}C$, with input images of $2048{\times}1024{\times}3$ pixels.
\label{fig:policynet}}
\end{figure*}

The policy network uses a lightweight 8-layer ResNet backbone combined with a fully convolutional head, with the architecture depicted in Figure~\ref{fig:policynet}. The backbone  operates on a feature representation $\mathcal{S}_t$ consisting of 4 inputs:
\begin{itemize}[noitemsep, topsep=0pt]
    \item Current frame $I_t$: the RGB frame at time $t$.
     \item Previous frame state $\mathcal{H}_{t-1}$: the previous frame state is an RGB frame, where each block has the image content of the last executed block for that position. By using the previous state, instead of simply the previous frame, we ensure that the network can detect small accumulating changes.
    \item Previous output $\mathcal{O}_{t-1}$: We represent detections or instances using a probability mask for each individual class. For segmentation, the output probabilities per pixel are used.
    \item Previous execution grid $\mathcal{A}_{t-1}$: The previous execution grid is a binary mask indicating which blocks were executed for the previous frame. In combination with the previous output, this improves the exploration ability of the policy, as previously executed blocks with no information gain are less likely to contain new information.

\end{itemize}

\subsection{Information Gain}
To determine the importance of each region, we define the Information Gain (IG) for each output pixel, as a quantity representing the amount of additional information gained by executing the model for that pixel compared to using the previous output. The information gain $IG_t$ at time $t$ is a function of the output $\mathcal{O}_t$ and the previous output $\mathcal{O}_{t-1}$.

The formulation of information gain is task-dependent, exploiting a task's characteristics to minimize the number of required computations while maximizing the output accuracy. The information gain is determined per pixel $p$, and combined afterwards per block $b$ using max-pooling:
\begin{equation}
    IG_b = \max IG_p \quad \forall p \in b ~.
\end{equation}

\paragraph{Object detection}
For object detection tasks, the information gain depends on the movement of objects and the score of the prediction. For every new frame, predicted bounding boxes are matched with the previous detections, by choosing the most overlapping detection using the intersection-over-union $IoU_{bb}$ of the bounding boxes. High overlap means low information gain, with static objects having no information gain. If an object does not overlap with any object detected in the previous frame, the object is a new detection and has an information gain equal to the detection score. Objects in the previous frame not matched with any object in the current frame also have information gain equal to the score of the previous detection, as those detections should be removed. 

Algorithm~\ref{alg:od_ig} is used to determine the information gain. Note that it is important to also assign information gain to pixels of the detections in the previous frame, in order to update or remove those detections when needed. Figure~\ref{fig:video} contains visualizations for the information gain.

\begin{algorithm}[tb!]
    \small
    \begin{algorithmic}
    
    \State \textbf{require:}~outputs $\mathcal{O}_{t}$, previous outputs $\mathcal{O}_{t-1}$
    \LineComment Initialize information gain as zero-filled matrix of size $H \times W$
    \State $IG \gets 0^{H{\times}W}$

    \ForAll {detections ${det}$ \textbf{in} $\mathcal{O}_{t}$ }
        \LineComment find most overlapping detection of previous output
        \State ${IoU}_{best} \gets 0$
        \State ${prevDet}_{best}  \gets \texttt{NULL}$
        \ForAll {detections ${prevDet}$ \textbf{in} $\mathcal{O}_{t-1}$ }
            \If{IoU($det$, $prevDet$) $> \text{IoU}_{best}$}
                \State ${IoU}_{best} \gets \text{IoU}(det, prevDet)$
                \State ${prevDet}_{best}  \gets {prevDet}$
            \EndIf
        \EndFor

        \LineComment set IG for pixels in bounding box of detection

        \ForAll{$\text{pixels}~p \in {det}$}
        \State $IG_p \gets \max(IG_p, (1 - IoU_{best})\cdot det_{score})$ 
        \EndFor
        \LineComment set IG for pixels in bounding box of matched detection

        \ForAll{$\text{pixels}~p \in {{prevDet}_{best}}$}
        \State $IG_p \gets \max(IG_p, (1 - IoU_{best}) \cdot prevDet_{score})$ 
        \EndFor

    \EndFor
    \LineComment process previous detections not overlapping with current ones 
    \ForAll {detections ${prevDet}$ \textbf{in} $\mathcal{O}_{t-1}$ }
        \If{$prevDet$ not processed}
            \State $IG_p \gets prevDet_{score} \quad \forall~\text{pixels}~p \in{prevDet}$
        \EndIf
        
    \EndFor
    
    \State \textbf{return} IG
    \end{algorithmic}
    \caption{Information Gain for Object Detection\label{alg:od_ig}}
\end{algorithm}

\paragraph{Instance segmentation}

The definition of information gain for instance segmentation is similar to the one of object detection, but the Intersection-over-Union is determined by the instance masks instead of the bounding boxes.

\paragraph{Semantic segmentation}
Semantic segmentation is a dense pixelwise classification task, where the network outputs a probability distribution per pixel. The information gain of each output pixel is determined by the pixelwise KL-Divergence between the output probability distributions. 

\subsection{Reinforcement learning}
The policy network $f_{pn}$ with parameters $\theta$ outputs a probability $p_b$ for each block $b$, indicating whether the features in that block should be calculated instead of just transferred. The network operates on the feature representation
\begin{equation}
    \mathcal{S}_t = \{I_t, \mathcal{H}_{t-1},  \mathcal{A}_{t-1},  \mathcal{O}_{t-1}\}
\end{equation}
 and outputs execution probabilities for each block $b$:
\begin{gather}
    \mathcal{P}_t = f_{pn} ( \mathcal{S}_t ; \theta )\\
\text{with } \mathcal{P}_t = [p_1, \dots, p_b, \dotsc, p_B] \in [0,1]^B.
\end{gather}
Probabilities $\mathcal{P}_t$ are sampled to execution decisions $\mathcal{A}_t =  [a_1, \dots, a_b, \dotsc, a_B] \in \{0,1\}^B$. The policy $\pi_{b,\theta}( {a}_b \mid \mathcal{S}_t)$ gives the probability of action $a_b$. Execution decision \mbox{$a_b=1$} results in execution of block $b$ and \mbox{$a_b=0$} results in feature transfer from the previous execution.

Stochastic sampling according to the probabilities encourages search space exploration, in comparison to simple thresholding. As gradients cannot be backpropagated through the sampling operation, we adopt reinforcement learning in order to optimize the policy for the task at hand.

Actions should maximize the reward for each block, with the objective to be maximized at each time step given by
\begin{equation}
\max \mathcal{J}(\theta) =\max  \sum_{b=1}^B{ \Bigl( \mathds{E}_{ a_b \sim \pi_{b,\theta}} \bigl[ \mathcal{R}_b(a_b) \bigr] \Bigl)}
\end{equation}
where $\mathcal{R}_b$ is the reward based on the Information Gain $IG$ as described later. The reward, loss, objective and parameters are determined at every timestep $t$, which we omit for simplicity of notation. 
The policy network's parameters $\theta$ can then be updated using gradient ascent with \mbox{learning rate $\alpha$}:
\begin{equation}
\theta \leftarrow \theta  + \alpha \nabla_{\theta} [\mathcal{J}(\theta)]~.
\end{equation}

Based on REINFORCE policy gradients~\cite{williams1992simple}, we can derive the loss function as (see supplemental material)
\begin{equation}
\mathcal{L}=  -{ \sum_{b=1}^{B} \bigl(\mathcal{R}_b(a_b) {  \log  \pi_{b,\theta}( a_b \mid \mathcal{S}_t) }} \bigr)~.
\end{equation}
The reward $\mathcal{R}_b$ depends on the information gain of that block. A trivial state would be to execute all blocks. Therefore, we introduce a reward $\mathcal{R}_{cost}$ weighted by hyperparameter $\gamma$ to balance the number of computations:
\begin{equation}
    {\mathcal{R}_b}(a_b) = {\mathcal{R}_{IG}}(a_b) +  \gamma \mathcal{R}_{cost}(a_b) ~.
\end{equation}
Executed blocks have a positive reward for positive information gain. In contrast, non-executed blocks have a negative reward to increase the likelihood of being executed:
\begin{equation}
    \mathcal{R}_{IG}(a_b) = 
    \begin{cases}
     IG_b &\text{ if } a_b = 1~, \\
      -IG_b &\text{ if } a_b = 0~.
    \end{cases}
    \label{eq:reward_ig}
\end{equation}
with ${IG}_b$ the information gain in a block.

The cost of a frame is the percentage of executed blocks:
\begin{equation}
\mathcal{C}_t =  \frac{\sum^B_i{a_i}}{B} \in [0,1] ~.
\end{equation}
As some frames might require more executed blocks than others, we define a moving average with momentum $\mu$:
\begin{equation}
    \mathcal{M}_t = (1-\mu) \cdot \mathcal{C}_{t}  + \mu \cdot \mathcal{C}_{t - 1} ~.
\end{equation}
Instead of simply minimizing the cost, we use a target parameter $\tau \in [0,1]$, which defines the desired average cost. This results in more stable training with less dependence on the exact value of $\gamma$. The cost reward is then given by
\begin{equation}
    \mathcal{R}_{cost}(a_b) = 
    \begin{cases}
    
     \tau - \mathcal{M}_t &\text{ if } a_b = 1~, \\
      - (\tau - \mathcal{M}_t) &\text{ if } a_b = 0~.
    \end{cases}
    \label{eq:reward_complexity}
\end{equation}
Executed blocks, where $a_b{=}1$, have a positive reward when the number of computations is lower than target $\tau$. The target could be adjusted at execution time, changing the model complexity on-the-fly.

\section{Experiments}

The acceleration achieved by BlockCopy is evaluated on pedestrian detection, instance segmentation and semantic segmentation tasks using datasets containing high-resolution video sequences. Pedestrian detection is a single-class 
object detection problem.
Our method is particularly suited for this task, 
as small persons need to be detected in high-resolution images. For each task, we integrate BlockCopy in state of the art existing networks, using publicly available implementations. Implementation details are given in the respective subsections.
In fully convolutional single-stage architectures, e.g. CSP~\cite{liu2019high_csp} and SwiftNet~\cite{orsic_defense_2019_swiftnet}, we integrate our block-sparse convolution with temporal feature propagation in all convolutional layers. For two-stage architectures, such as Mask-RCNN~\cite{he2017mask} and MGAN~\cite{pang2019mask_mgan}, only the backbone and region proposal network are optimized, with the network head applied as normal.

\paragraph{Datasets}
The Cityscapes~\cite{cityscapes} dataset is used to evaluate instance and semantic segmentation. The dataset consists of 2975, 500 and 1525 video sequences for training, validation and testing respectively. Each video sequence contains 20 frames of $2048{\times}1024$ pixels recorded at 17 Hz, with the last frame having detailed semantic and instance annotations. We use the standard 19 classes for semantic segmentation and 8 classes for instance segmentation. CityPersons~\cite{citypersons} builds on Cityscapes by adding high-quality bounding box annotations for 35000 persons.

Note that we do not use ground-truth annotations to train our method, as the policy network is trained in a self-distilled fashion. Ground-truth labels are only used to evaluate accuracy after speeding up inference. 

\paragraph{Evaluation Metrics} The accuracy of pedestrian detection is evaluated using the log-average miss rate (MR) criterion, following the standard metrics (Reasonable, Bare, Partial, Heavy) of CityPersons~\cite{citypersons}.
Instance segmentation is evaluated using COCO-style mean average precision (AP) and AP50 for an overlap of 50\%, while semantic segmentation uses the mean Intersection-over-Union (mIoU) metric.

Computational cost is measured by two metrics: the number of operations and the inference speed. The number of operations is reported as billions of multiply-accumulates (GMACS). 
Inference speed is measured as the average processing time per frame, including data loading and post-processing, on an Nvidia GTX 1080 Ti 11 GB GPU with an Intel i7 CPU, PyTorch 1.7 and CUDA 11. 

\paragraph{BlockCopy configuration}
For all tasks, we train the policy network in an online fashion using an RMS optimizer with learning rate $1e^{-4}$ and weight decay $1e^{-3}$. To mitigate 
the impact of the backward pass and weight updates, we only update the policy weights every 4 frames. Before running on the validation or test set, the policy is initialized on 400 training clips. The first frame of a video sequence is always executed completely, and the following 19 frames are processed sparsely using the BlockCopy framework and policy. We evaluate each network and task with varying cost targets $\tau \in [0,1]$ in order to obtain models with different computational complexities. Hyperparameter $\gamma$ setting the balance between the reward terms is fixed to 5, with the cost momentum $\mu$ set to 0.9.

\paragraph{Baselines}
Besides comparisons with other methods, 
we compare our inference speedup method with lower spatial resolution and lower frame rate baselines. Reducing the 
input resolution decreases the number of operations and increases inference speed, with worse predictions for small objects. Decreasing the frame rate by skipping frames decreases temporal resolution.
Our experiments show that lowering the frame rate has a significant impact on accuracy, underlining the important of fast processing. 

\subsection{Pedestrian detection}

We integrate our method in the Center and Scale Predictor (CSP)~\cite{liu2019high_csp} and Mask-Guided Attention Network (MGAN)~\cite{pang2019mask_mgan} architectures. CSP is a single-stage anchor-free detector, predicting the center and scale of each object with the aspect ratio fixed to $0.41$. It builds on the \mbox{ResNet-50}~\cite{he2016deep_resnet} backbone.
MGAN is a dual-stage detector using the VGG~\cite{simonyan2014very_vgg} backbone. Our implementation is based on the Pedestron~\cite{hasan2020generalizable_pedestron} framework.
The standard evaluation setting uses 17 Hz video sequences at $2048{\times}1024$ resolution on the Citypersons dataset~\cite{citypersons}. 

\begin{figure}[!tb]
\vspace*{-5mm}
\centering
\subfloat[Miss Rate vs. GMACS]{\includegraphics[width=0.5\linewidth]{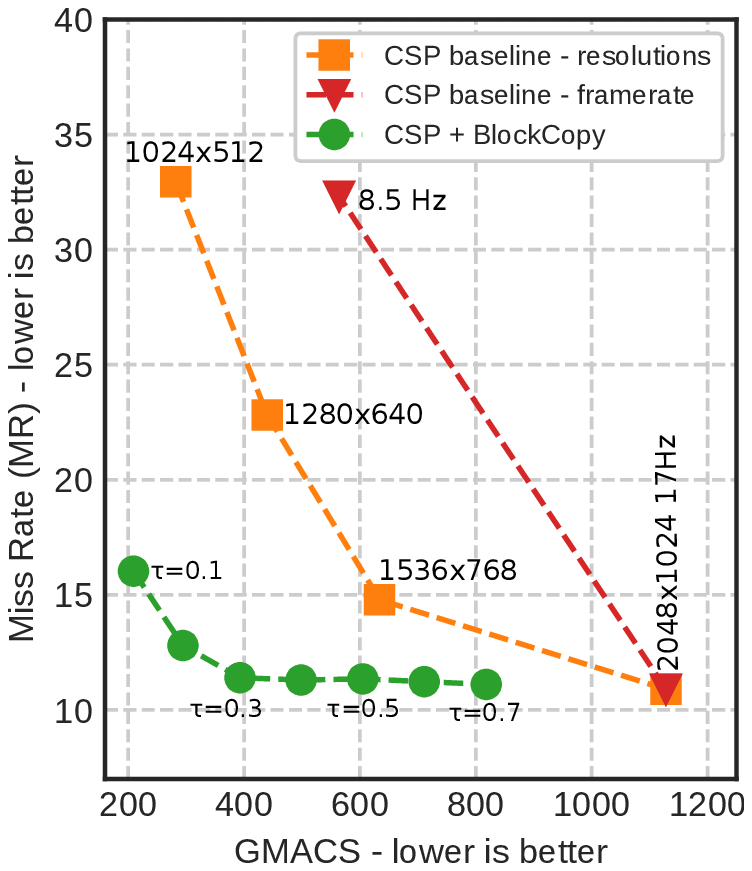}%

\label{fig:citypersons_gmacs}}
\hfil
\subfloat[Miss Rate vs. inference time]{\includegraphics[width=0.5\linewidth]{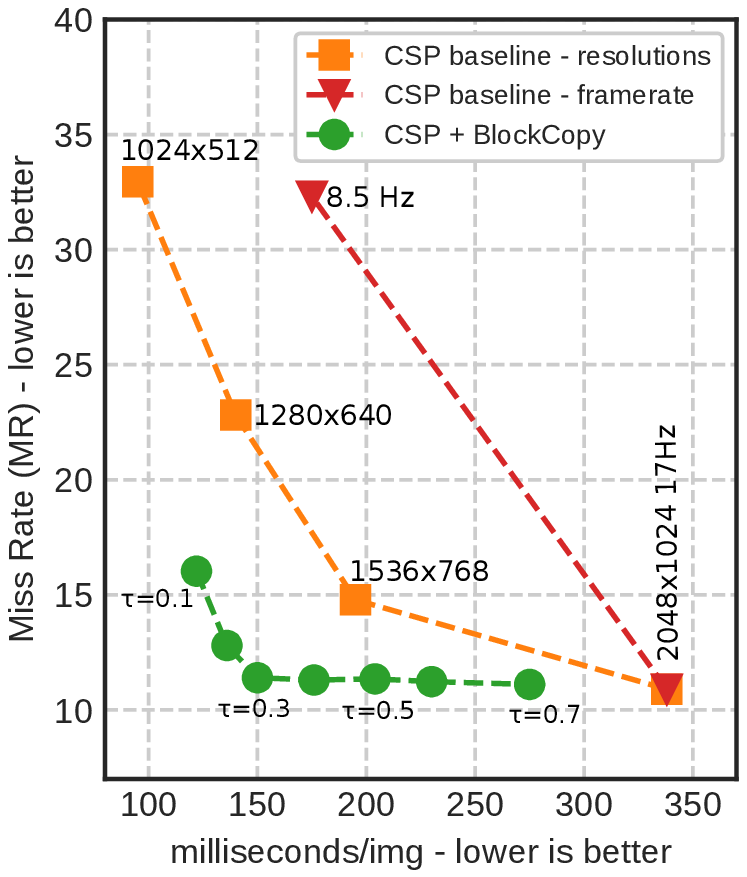}%
\label{fig:citypersons_fps}}
\vspace*{-2mm}
\caption{Results on CityPersons validation set on CSP~\cite{liu2019high_csp} with BlockCopy. Miss Rate (MR) is reported for the Reasonable subset. Models with BlockCopy consistently outperform lower resolution baselines of similar complexity.}
\label{fig:citypersons_val}
\end{figure}

The detection results for CSP~\cite{liu2019high_csp} with BlockCopy are shown in Figure~\ref{fig:citypersons_gmacs} and Figure~\ref{fig:citypersons_fps}, when comparing the number of operations (GMACS) and inference time respectively. 
BlockCopy models (with different target costs $\tau$) achieve better detection results (lower miss rate) than lower resolution and lower frame rate baselines, demonstrating improved efficiency.  With $\tau{=}0.3$, the amount of operations and processing time is more than halved with only 0.4\% increase in miss rate.  Table~\ref{tab:ped_val} compares BlockCopy to other methods and demonstrates that our method is faster than existing methods while achieving competitive accuracy.

\begin{table*}[]
\small
\setlength\tabcolsep{4pt} 
\centering
\caption{Results (log-average Miss Rate) on CityPersons val/test set. Values marked with~\dag~were determined using the Pedestron~\cite{hasan2020generalizable_pedestron} implementation. Inference time per frame measured on a GTX 1080 Ti GPU.}
\vspace*{-2mm}
\label{tab:ped_val}
\begin{tabular}{@{}l|l|r|r|rrrH|rr@{}}
\toprule
Method                                                             & Backbone                      & Reasonable [test] & Reasonable& Bare& Partial& Heavy                    & Small                    & GMACs    & avg. sec/img \\ \midrule
{CSP + BlockCopy ($\tau{=}0.3$)}                 & {ResNet-50}& 12.50 &11.4    &7.6    &    10.8 & {49.5    }& {    }&    393      & 0.151 s         \\
{MGAN + BlockCopy ($\tau{=}0.3$)}                & {VGG-16}   & 10.83 &11.2   & 6.3   &    10.9 & {60.5    }& {15.35    }&   560       & 0.140 s\\ \midrule
{CSP (CVPR2019)~\cite{liu2019high_csp}}          & {ResNet-50}& - &11.0& 7.3& 10.4& {49.3}& {16.0}& 1128 \dag& 0.330 s                                    \\
{MGAN (ICCV2019)~\cite{pang2019mask_mgan}}      & {VGG-16}   & - &11.0& -  & -   & {50.3}& { -  }& 1104 \dag& 0.224 s \dag                             \\
{MGAN scale $\times1.3$ ~\cite{pang2019mask_mgan}}      & {VGG-16}   & 9.29 & 9.9 & -  & -   & 45.4 & { -  }& 1665 \dag & 0.370 s \dag                            \\
{ALFNet (ECCV2018)~\cite{liu2018learning_alfnet}}& {ResNet-50}&- &12.0& 8.4& 11.4& {51.9}& {19.0}& -        & 0.270 s\\
{AdaptiveNMS (CVPR2019)~\cite{liu2019adaptive}}& {VGG-16}& 11.17 &10.8& 6.2&11.4 &54.0&- &    -    &- \\
{OR-CNN (ECCV2018)~\cite{zhang2018occlusion_orcnn}}& {VGG-16}& 11.32 &11.0& 5.9&13.7 &51.3&- & -       &- \\
{HBAN (Neurocomputing2020)~\cite{lu2020semantic_hban}}& {VGG-16}& 11.26 &10.7&- &- & 46.9& &     -  &  0.760 s\\

\bottomrule      
\end{tabular}
\end{table*}

\subsection{Instance segmentation}
\begin{figure}[!tb]
\vspace*{-8mm}
\centering
\subfloat[AP vs. GMACS]{\includegraphics[width=0.5\linewidth]{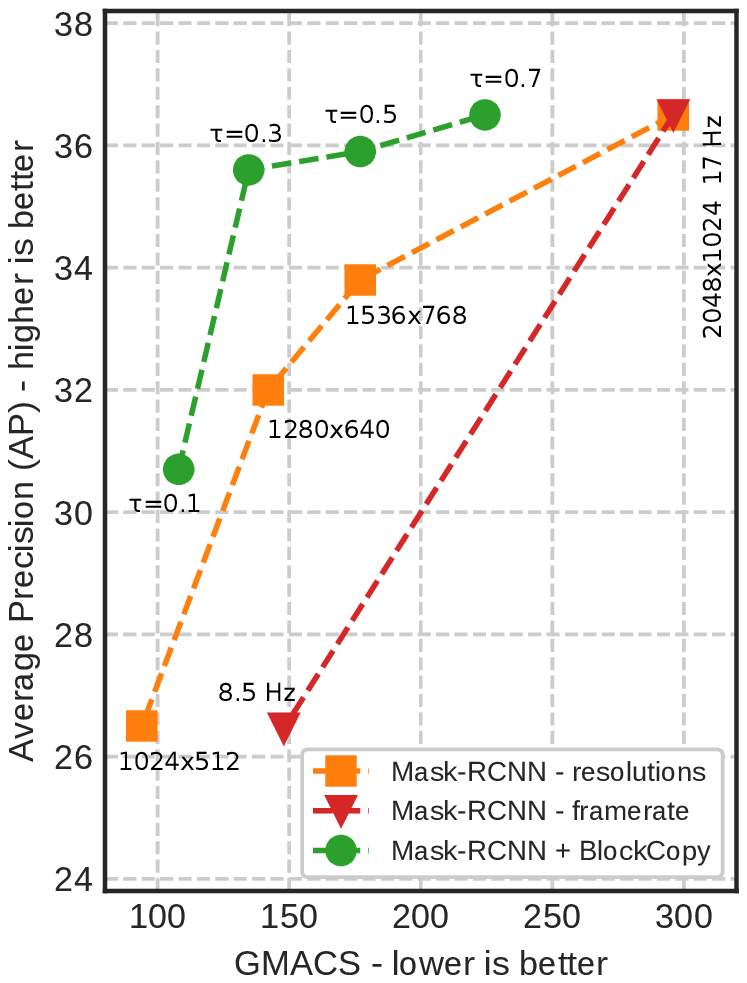}%
\label{fig:cityinstance_gmacs}}
\hfil
\subfloat[AP vs. inference time]{\includegraphics[width=0.5\linewidth]{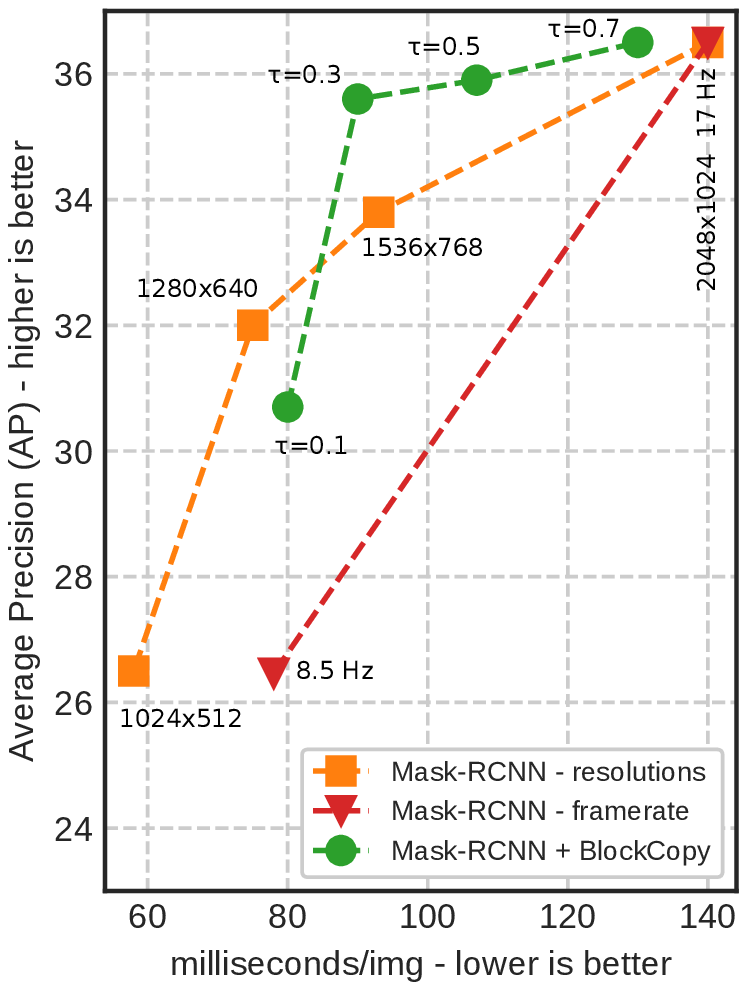}%
\label{fig:cityinstance_fps}}
\vspace*{-3mm}
\caption{Cityscapes instance segmentation \textit{val} set.}
\label{fig:cityinstance_val}
\end{figure}

We integrate BlockCopy in the popular Mask-RCNN~\cite{he2017mask} architecture with  ResNet-50~\cite{he2016deep_resnet} backbone for instance segmentation, using the baseline provided by Detectron2~\cite{wu2019detectron2}.
Figure~\ref{fig:cityinstance_gmacs}~and~\ref{fig:cityinstance_fps} show that integrating BlockCopy with $\tau{=}0.3$ halves the amount floating point operations with 0.9\% accuracy decrease, while the frames processed per second increases from 6.7 to 11.0 FPS. The test set submission of our method (with $\tau{=}0.5$) achieved 31.7 AP, compared to 32.0 AP of the Mask-RCNN baseline~\cite{he2017mask}, with ${\times}1.65$ faster inference.

\subsection{Semantic segmentation}
We compare BlockCopy with other video processing using optical flow and adaptive keyframe scheduling on a semantic segmentation task. Our method can be seen as a sparse version of adaptive scheduling, where each block is scheduled individually.

BlockCopy is integrated in the popular DeepLabV3+~\cite{chen2018encoder_deeplab} network combined with a ResNet-101~\cite{he2016deep_resnet} backbone, and the faster SwiftNet~\cite{orsic_defense_2019_swiftnet} model with ResNet-50.

Since other methods report inference time on various GPUs, we scaled those values to be equivalent to a Nvidia GTX 1080 Ti, as proposed by other works for fair comparison~\cite{orsic_defense_2019_swiftnet}.
All data, including non-compensated inference time and the GPU scaling factors, can be found in supplemental. Figure~\ref{fig:semseg} shows that our method is competitive with methods designed specifically for semantic segmentation and achieves higher mIoU with lower inference time.

\begin{figure}[tb!]
\centering
\includegraphics[width=0.9\linewidth]{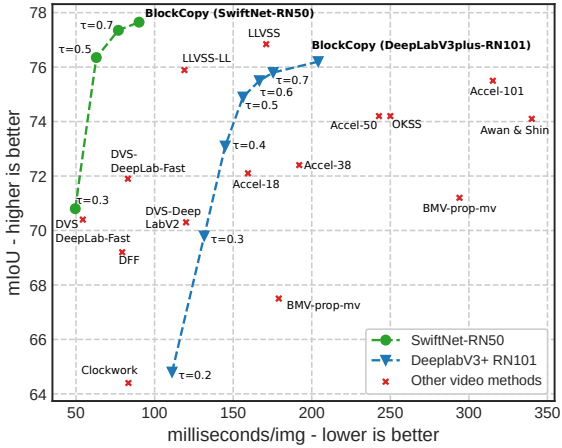}
\vspace*{-3mm}
\caption{Results on Cityscapes semantic segmentation validation set. Inference time of other methods is compensated for GPU performance to match GTX 1080 Ti (see supplemental). Other video methods are Accel~\cite{jain2019accel}, Awan and Shin~\cite{awan2020warping}, BMV-prop-mv~\cite{jain2018fast_bmv}, Clockwork~\cite{shelhamer2016clockwork}, DFF~\cite{zhu2017deep_dff}, DVSNet~\cite{xu2018dynamic_dvs}, LLVSS~\cite{li2018low_llvss} and  OKSS~\cite{awan2020online}.
\label{fig:semseg}}
\end{figure}

\section{Ablation}
Ablation for the policy network is given in Table~\ref{tab:policy_ablation} and shows that including more information about the previous frame is beneficial. Online learning slightly outperforms offline learning. The overhead introduced by BlockCopy is given in Table~\ref{tab:ablation_overhead}. The execution of the policy network and the updates of the weights based on information gain are relatively cheap compared to the task network.

\begin{table}[tb]
\setlength\tabcolsep{2pt} 
\footnotesize
\caption{Ablation for policy network. }
\vspace*{-3mm}
\label{tab:policy_ablation}
\begin{tabular}{@{}l|c|ccccc|c|c@{}}
\toprule
Backbone                           & Online & $\mathcal{I}_{t}     $& $\mathcal{I}_{t-1}$    & $\mathcal{F}_{t-1}$    & $\mathcal{O}_{t-1}$    & $\mathcal{A}_{t-1}$   & GMACS & MR \\ \midrule
\multirow{6}{*}{
ResNet-8
}
& \checkmark & \checkmark &            &            &            &            & 6.4 & 13.0 \%   \\
                                & \checkmark & \checkmark & \checkmark &            &            &            &  6.4 & 12.0 \% \\
                               & \checkmark & \checkmark &           & \checkmark &            &           & 6.4  & 12.0 \% \\
                               &\checkmark  & \checkmark &            & \checkmark & \checkmark &            & 6.5  & 11.7 \% \\
                                & \checkmark & \checkmark & \checkmark           & \checkmark           &            & \checkmark           & 6.5 &  11.5\%   \\
                               &\checkmark  & \checkmark &            & \checkmark & \checkmark & \checkmark &  6.5  & 11.4 \% \\
                               &  & \checkmark &            & \checkmark & \checkmark & \checkmark & 6.5 & 13.3 \%  \\ \hdashline
 ResNet-20                & \checkmark  & \checkmark &            & \checkmark & \checkmark & \checkmark &  34.1 & 11.5 \% \\
                                 \bottomrule
\end{tabular}
\end{table}

\begin{table*}[tb!]
\centering
\setlength\tabcolsep{3pt} 
\footnotesize

\caption{Overhead of BlockCopy components in the CSP network for pedestrian detection~\cite{liu2019high_csp}. 
}
\label{tab:ablation_overhead}
\vspace*{-3mm}
\begin{tabular}{@{}l|c|cccH|cccH|cc|c@{}}
\toprule
\textbf{Method}                               & \textbf{Total}& \multicolumn{4}{c|}{\textbf{Task network}}& \multicolumn{4}{c|}{\textbf{Policy}}& \textbf{\makecell[c]{GMACS Task} }& \textbf{\makecell[c]{GMACS Policy} }& \textbf{Acc. (MR)} \\ \midrule
                                              &       & \makecell[c]{Sparse Conv.\\ Overhead}& \makecell[c]{Feature\\Transfer}& Ops   & Total& \makecell[c]{Policy\\Network}& \makecell[c]{Inform.\\Gain}& \makecell[c]{Backward pass + \\ Weight update}& Total&     &     & \\ \midrule
CSP baseline                                  & 330 ms& N.A.                                 & N.A.                           & 330   & 330  & N.A.                         & N.A.                       & N.A.                                          & N.A. & 1128& N.A.& 11.0 \% \\ \hdashline
\makecell[l]{CSP + BlockCopy\\ ($\tau$ = 0.7)}& 275 ms& 30 ms                                & 9 ms                           & 215 ms& -    & 9 ms                         & 3 ms                       & 9 ms                                          & -    & 812 (-28\%) & 6.5 & 11.1 \%\\
\makecell[l]{CSP + BlockCopy\\ ($\tau$ = 0.5)}& 204 ms& 25 ms                                & 12 ms                          & 146 ms& -    & 9 ms                         & 4 ms                       & 8 ms                                          & -    & 599 (-47\%) & 6.5 & 11.3 \%\\
\makecell[l]{CSP + BlockCopy\\ ($\tau$ = 0.3)}& 150 ms& 21 ms                                & 15 ms                          & 92 ms & -    & 9 ms                         & 3 ms                      & 10 ms                                          & 26 ms& 388 (-65\%) & 6.5 & 11.4 \% \\
\bottomrule
\end{tabular}
\end{table*}

\section{Conclusion}
We proposed the BlockCopy framework that can be applied to existing pre-trained convolutional neural networks, improving their efficiency for high-resolution video-based processing. We integrated the method in a variety of networks for different computer vision tasks, demonstrating strong inference speedup with only a small drop in accuracy. Tasks such as pedestrian detection and instance segmentation are particularly suited for this method, as only few image areas are important. By not requiring training labels, our method can be integrated in deployment pipelines starting from existing pre-trained models. 

\section*{Acknowledgement}
This work was funded by FWO on the SBO project with agreement S004418N.
{\small
\bibliographystyle{ieee_fullname}
\bibliography{egbib}
}

\end{document}